\documentclass{article}

\usepackage[preprint]{neurips_2026}

\usepackage[utf8]{inputenc}
\usepackage[T1]{fontenc}
\usepackage{hyperref}
\usepackage{url}
\usepackage{booktabs}
\usepackage{amsmath,amssymb,amsfonts,amsthm,mathtools}
\usepackage{nicefrac}
\usepackage{microtype}
\usepackage{xcolor}
\usepackage{enumitem}
\usepackage{needspace}
\usepackage{placeins}

\hypersetup{hidelinks}

\theoremstyle{plain}
\newtheorem{theorem}{Theorem}[section]
\newtheorem{proposition}[theorem]{Proposition}
\newtheorem{lemma}[theorem]{Lemma}
\newtheorem{corollary}[theorem]{Corollary}
\theoremstyle{definition}
\newtheorem{definition}[theorem]{Definition}
\newtheorem{example}[theorem]{Example}
\theoremstyle{remark}
\newtheorem{remark}[theorem]{Remark}

\newcommand{\R}{\mathbb{R}}
\newcommand{\N}{\mathbb{N}}

\newcommand{\KL}{\operatorname{KL}}
\newcommand{\rank}{\operatorname{rank}}

\newcommand{\diag}{\operatorname{diag}}
\newcommand{\softmax}{\operatorname{softmax}}
\newcommand{\im}{\operatorname{im}}
\newcommand{\Span}{\operatorname{span}}
\newcommand{\dd}{\,\mathrm{d}}
\newcommand{\cX}{\mathcal X}
\newcommand{\cY}{\mathcal Y}
\newcommand{\cL}{\mathcal L}

\newcommand{\one}{\mathbf 1}
\newcommand{\quot}[1]{\overline{#1}}

\title{Same Function, Different Mechanism: Implementation-Local Geometry Beyond Accuracy}

\author{Miroslav L\v{z}i\v{c}a\v{r}\\
Mereology Labs\\
\texttt{miroslav.lzicar@mereologylabs.com}}

\begin{document}
\maketitle

\begin{abstract}
Two models can make exactly the same predictions yet respond differently when their implementations are perturbed. A lookup table can change one case independently; a shared rule couples changes across cases. We compare these mechanisms through an \emph{implementation chart}: a specified local family of logit perturbations around a fixed predictive distribution. Our main result counts the independent prediction changes allowed by exact finite-group sharing: the active dimension is the number of input-output orbits minus the number of input orbits. For softened modular addition, lookup and shared cyclic charts therefore have ranks $p^2(p-1)$ and $p-1$ despite identical predictions. The classical Fisher pullback connects these dimensions to local KL curvature; mechanism-preserving reparameterizations leave the rank unchanged. We also bound the spectral tail under approximate sharing and separate quadratic rank from higher-order learning coefficients. Trained modular-addition models verify that chart rank is measurable, while controlled compiled transformers distinguish compiler-supported intermediate charts from matched alternatives. Negative controls show where this discrimination fails. The framework evaluates specified mechanism hypotheses; it does not infer a unique mechanism from predictions or predict training dynamics.
\end{abstract}

\section{Introduction}

Suppose two systems answer every modular-addition question correctly, with exactly the same confidence. One stores a separate answer for each input pair. The other reuses one cyclic rule. Their predictions cannot distinguish them, but an intervention can: changing one stored answer need not affect another, whereas changing the shared rule affects many inputs together. This paper asks how to measure that difference while holding the predictive distribution fixed.

The key quantity is the number of \emph{independent local changes to predictions} that a specified mechanism permits. We call this its \emph{active dimension}. To define it, we need more than the model's outputs: we must say which parts of the implementation can change and which constraints remain in place. An \emph{implementation chart} records that choice. For example, perturbing a convolutional kernel while preserving weight tying defines a different chart from independently perturbing its copies at every position. Both charts can pass through the same predictive distribution.

\paragraph{A running example.}
Let $a,b,y\in\mathbb Z_p$. Think of the output logits as a cyclic array indexed by $y$. A shared implementation reads one kernel at the shifted position $y-a-b$; an unrestricted lookup implementation stores a separate array for each $(a,b)$. Adding the same constant to all logits does not change softmax probabilities, so each array has $p-1$ meaningful directions. Lookup permits $p^2(p-1)$ independent changes; sharing permits only $p-1$. Sections~\ref{sec:setup}--\ref{sec:modular} formalize this example and explain why the gap survives coordinate changes and redundant parameters.

\Needspace{8\baselineskip}
\paragraph{Main contributions.}
\begin{itemize}[leftmargin=*,itemsep=2pt]
\item \textbf{An orbit-count formula for shared mechanisms.} Theorem~\ref{thm:group-sharing} computes active dimension from group orbits, without a coordinate-level Hessian calculation. Modular addition gives an unbounded factor-$p^2$ gap at the same predictive distribution.
\item \textbf{A precise comparison of local mechanism hypotheses.} Charts with the same tangent image represent the same first-order perturbation family. Reparameterization, dead coordinates, and duplication preserve its rank (Proposition~\ref{prop:invariance}); changing the allowed family changes the question. The Fisher pullback used to measure rank is classical background (Lemma~\ref{thm:pullback}).
\item \textbf{An operational test and its boundaries.} We compute chart ranks on trained networks and compare intervention-defined charts in compiled transformers (Section~\ref{sec:operational}). A spectral-tail bound handles approximate sharing, and explicit higher-order examples show why rank alone is not a general learning-coefficient estimator.
\end{itemize}

\paragraph{Why local geometry?}
Singular learning theory (SLT) associates Bayesian complexity with a local statistical model, not just its predictions \citep{watanabe2009,watanabe2013,watanabe2018}. Work on neural local learning coefficients (LLCs) and program singularities motivates comparing implementations at a common predictive distribution \citep{wei2023deep,lau2025llc,hoogland2024developmental,wang2024rllc,murfetTroiani2025programs}. We isolate the part visible to second-order geometry and state where higher-order terms matter. Neither an unrestricted Hessian nor the results below determine which algorithm a network learned, or predict grokking and other training transitions \citep{power2022grokking,nanda2023progress}. Chart specification requires architectural, compiler, or intervention evidence.

\section{Finite-output mechanisms and logit geometry}
\label{sec:setup}

We first fix the predictions, then specify their allowed perturbations.

Let $\cX$ be a finite input domain with $\mu(x)>0$ and let $\cY=\{1,\ldots,k\}$ be a finite label set. Let $q_x\in\Delta^{k-1}$ be a full-support target conditional distribution for each $x$. Logits are identifiable only up to additive constants, so define the quotient logit space
\begin{equation}
  \quot{\R^k}=\R^k / \Span\{\one\},
  \qquad
  \cL = \bigoplus_{x\in\cX}\quot{\R^k}.
\end{equation}
The quotient removes one unobservable offset per input. In the running example, it leaves $p-1$ logit contrasts for each of the $p^2$ input pairs.

The softmax map is well-defined on $\quot{\R^k}$. Write $z_x^*=[\log q_x]\in\quot{\R^k}$ and $z^*=(z_x^*)_{x\in\cX}\in\cL$. For $z\in\cL$, define the population KL loss
\begin{equation}
  K(z) = \sum_{x\in\cX}\mu(x)\,\KL\!\left(q_x\middle\|\softmax(z_x)\right).
  \label{eq:ambient-kl}
\end{equation}
Then $K(z^*)=0$ and $z^*$ represents the predictive distribution, not a particular neural parameterization.

\begin{definition}[Implementation chart]
\label{def:implementation-chart}
An \emph{implementation chart} is a $C^2$ map $\chi_A:U_A\subseteq\R^{d_A}\to\cL$ with $0\in U_A$ and $\chi_A(0)=z^*$. The index $A$ denotes the intended mechanism: for example, independent lookup entries, a shared rule, a low-rank factorization, a circuit subspace, or a compiler-specified program. The resulting local loss (its behavior in a neighborhood of $u=0$, also called a germ) is
\begin{equation}
  K_A(u)=K(\chi_A(u)).
\end{equation}
Two charts are extensionally equivalent at the base point if they have the same $z^*$; they may still have different local geometry because their perturbation families differ.
\end{definition}

For Hessian-rank statements, the $C^2$ assumption is sufficient. Whenever we discuss an actual real log canonical threshold or local learning coefficient, we additionally assume the relevant chart germ is analytic and the prior density is positive and smooth in a neighborhood of the minimizing set.

For the cyclic example, the lookup coordinates change the input-specific arrays independently, while shared coordinates change the common kernel. At $u=0$ both give the same $q$.

This definition intentionally separates the predictive distribution from the local family of alternatives. A neural network restricted to a finite evaluation domain induces such a chart by composing parameter perturbations with the logits on that domain. A hand-written algorithmic or architectural mechanism induces a chart directly by specifying which perturbations are allowed.

\section{Measuring active dimension: the Fisher pullback}
\label{sec:pullback}

For each $x$, define the Fisher covariance on logits
\begin{equation}
  C_x = \diag(q_x)-q_xq_x^\top .
\end{equation}
It annihilates the gauge direction $\one$ and is positive definite on $\quot{\R^k}$. Let $F_q$ be the block-diagonal positive-definite bilinear form on $\cL$ given by
\begin{equation}
  F_q(v,w)=\sum_{x\in\cX}\mu(x)\, v_x^\top C_x w_x,
  \label{eq:fisher-metric}
\end{equation}
where representatives in $\R^k$ may be chosen arbitrarily because $C_x\one=0$.

\begin{lemma}[Classical logit-Fisher pullback]
\label{thm:pullback}
Let $\chi_A:U_A\to\cL$ be an implementation chart and let $A=D\chi_A(0):\R^{d_A}\to\cL$. Then
\begin{equation}
  \nabla K_A(0)=0,
  \qquad
  \nabla^2K_A(0)=A^\top F_q A .
  \label{eq:pullback-hessian}
\end{equation}
Consequently,
\begin{equation}
  \rank\nabla^2K_A(0)=\dim \im A .
  \label{eq:active-rank}
\end{equation}
On the active quotient $\R^{d_A}/\ker A$, the quadratic germ is regular and has quadratic active contribution
\begin{equation}
  \lambda_A^{\rm quad}=\frac12\dim\im A .
  \label{eq:regular-target}
\end{equation}
If $\chi_A$ is analytic and its tangent-kernel directions split off as genuinely dead local coordinates, this quadratic contribution is the local learning coefficient under a positive smooth prior. Restriction to an active transverse slice also gives $r/2$, but describes that restricted chart rather than the full germ. Otherwise Eq.~\eqref{eq:regular-target} is only the second-order active contribution; higher-order protected directions in $\ker A$ may change the actual learning coefficient.
\end{lemma}

\paragraph{How to read the lemma.}
The Jacobian $A$ converts parameter changes into logit changes. Multiplication by $F_q^{1/2}$ weights those changes by predictive uncertainty and input probability. Its rank counts independent local prediction changes; its squared singular values give the nonzero eigenvalues of the pullback. In the lookup example, different inputs can move independently. In the shared example, all inputs reuse the same directions. This is the standard Fisher chain rule applied to a specified mechanism chart, not a new identity.

\begin{proof}[Proof sketch]
For one input $x$, the gradient of $\KL(q_x\|\softmax(z_x))$ with respect to logits is $\softmax(z_x)-q_x$, which vanishes at $z_x^*$. The Hessian at $z_x^*$ is $C_x$. Summing over inputs gives the ambient Hessian $F_q$. The chain rule gives $A^\top F_q A$ because the first derivative of $K$ vanishes at $z^*$. Since $F_q$ is positive definite on the quotient logit space, the rank of the pullback is exactly $\dim\im A$. A rank-$r$ positive quadratic form has regular real log canonical threshold $r/2$ on its active coordinates. Appendix~\ref{app:pullback-proof} gives the quotient-space details.
\end{proof}

\begin{corollary}[No function-only active dimension]
\label{cor:no-function-only}
Fix any full-support finite conditional distribution $q$. For any linear subspace $S\subseteq\cL$, the affine chart $\chi_S(u)=z^*+u$ with $u\in S$ implements $q$ at $u=0$ and has active rank $\dim S$. Thus the local active dimension is not determined by the conditional distribution alone.
\end{corollary}

Corollary~\ref{cor:no-function-only} is the formal version of the slogan ``same function, different mechanism.'' It also explains why a chart condition is unavoidable: Bayesian evidence and SLT attach complexity to a local statistical model, not only to a point distribution.

\section{What chart dependence is allowed?}
\label{sec:chart-dependence}

The numerical curvature of a loss depends on coordinates. Its rank at a stationary point survives invertible smooth coordinate changes. Here we ask a broader question: which changes preserve the same family of first-order prediction changes?

\begin{definition}[First-order mechanism equivalence]
Two implementation charts $\chi_A$ and $\chi_B$ at the same $z^*$ are \emph{first-order mechanism equivalent} if their tangent images in $\cL$ agree:
\begin{equation}
  \im D\chi_A(0)=\im D\chi_B(0).
\end{equation}
They represent the same local perturbation mechanism to first order, although they may use different coordinates or redundant parameters.
\end{definition}

\begin{proposition}[Invariance under reparameterization, dead code, and duplication]
\label{prop:invariance}
Let $\rho:V\subseteq\R^{d'}\to U_A\subseteq\R^{d_A}$ be $C^2$, $\rho(0)=0$, and suppose $D\rho(0)$ is surjective. Then the composed chart $\chi_A\circ\rho$ has the same active rank as $\chi_A$:
\begin{equation}
  \rank \nabla^2(K_A\circ\rho)(0)=\rank \nabla^2K_A(0).
\end{equation}
This includes local diffeomorphisms, adding dead coordinates $(u,v)\mapsto u$, and duplicated coordinates such as $(u_1,u_2)\mapsto u_1+u_2$. If the charts and $\rho$ are analytic, the prior has positive smooth product density, and the local fibers have finite volume, then the pullback germ has the same learning coefficient as the original germ. In the dead and duplicated examples the extra coordinates split off as inert fibers after a local change of variables.
\end{proposition}

\begin{proof}[Proof sketch]
The tangent image of $\chi_A\circ\rho$ is $\im(D\chi_A(0)D\rho(0))=\im D\chi_A(0)$ by surjectivity. The rank statement follows from Lemma~\ref{thm:pullback}. Under the analytic and prior assumptions, the learning-coefficient statement follows from the constant-rank theorem: locally, the redundant coordinates split into active coordinates and inert fibers. See Appendix~\ref{app:invariance-proof}.
\end{proof}

\begin{remark}[Changing the perturbation class changes the question]
\label{prop:ambient}
Let $S\subsetneq T\subseteq\cL$ be two mechanism subspaces and consider the charts $z^*+S$ and $z^*+T$. They implement the same distribution at the base point but have active ranks $\dim S$ and $\dim T$. In particular, a shared implementation embedded as a point in a larger lookup model has lookup-rank local geometry if the prior permits arbitrary lookup perturbations, and shared-rank local geometry if the prior is concentrated on the shared submodel.
\end{remark}

Remark~\ref{prop:ambient} identifies a modeling distinction rather than a coordinate artifact. The posterior preference is over a local model class or prior support; the same parameter value can be evaluated under a broad ambient model or under a mechanism-preserving submodel.

\section{Shared mechanisms from finite-group structure}
\label{sec:group}

Many mechanisms reuse computation because a symmetry identifies perturbations across examples. An \emph{orbit} is a set of entries related by that symmetry. A shared logit perturbation assigns one value to each input-output orbit. Some of these values merely shift every output at an input together and are unobservable. Subtracting those input-orbit offsets gives the main result.

For the cyclic example, translating $(a,b)$ by $(s,t)$ and $y$ by $s+t$ preserves the residue $y-a-b$. Every input pair can be reached from every other, so there is one input orbit. Triples with the same residue form one orbit, giving $p$ input-output orbits. The count is therefore $p-1$.

Let a finite group $G$ act on $\cX$ and on $\cX\times\cY$ compatibly with the projection to $\cX$. Equivalently, $g$ sends an input $x$ to $gx$ and relabels outputs over that input. Assume $q$ is $G$-equivariant: $q_{gx}(gy)=q_x(y)$. The group acts linearly on raw logits $V=\R^{\cX\times\cY}$ and on the gauge subspace $B=\{b_{x,y}=a_x\}$, hence on $\cL=V/B$.

\begin{definition}[Lookup and $G$-shared charts]
The \emph{lookup chart} is the full quotient logit chart $z^*+\cL$. The \emph{$G$-shared chart} is the invariant affine subspace $z^*+\cL^G$, where $\cL^G$ denotes the $G$-fixed subspace.
\end{definition}

\Needspace{15\baselineskip}
\begin{theorem}[Orbit-count active dimension for exact sharing]
\label{thm:group-sharing}
For full-support $G$-equivariant $q$, the lookup chart has active rank
\begin{equation}
  r_{\rm lookup}=|\cX|(k-1).
\end{equation}
The $G$-shared chart has active rank
\begin{equation}
  r_G = \dim\cL^G
  = \#\big((\cX\times\cY)/G\big)-\#(\cX/G),
  \label{eq:orbit-rank}
\end{equation}
where $\#(S/G)$ is the number of $G$-orbits in $S$. Their quadratic active contributions are $r_{\rm lookup}/2$ and $r_G/2$, and these equal the LLCs of these affine charts by Lemma~\ref{thm:pullback}.
\end{theorem}

\begin{proof}[Proof sketch]
The lookup rank is $\dim\cL=|\cX|k-|\cX|$. For the shared chart, Lemma~\ref{thm:pullback} reduces the problem to computing $\dim\cL^G$. Since $G$ is finite and the field has characteristic zero, taking invariants is exact, so $(V/B)^G\cong V^G/B^G$. The dimension of $V^G$ is the number of orbits on input-output pairs, and the dimension of $B^G$ is the number of orbits on inputs. Appendix~\ref{app:group-proof} gives the full argument.
\end{proof}

Theorem~\ref{thm:group-sharing} connects local Bayesian complexity to weight sharing and equivariance \citep{cohen2016group,kondor2018generalization,bronstein2021geometric}. It is not a claim that all useful mechanisms are symmetries; rather, exact group sharing is a large class where the mechanism-local rank is computable without inspecting coordinates one by one.

\section{Modular addition: lookup versus a shared cyclic rule}
\label{sec:modular}

Let $\cX=\mathbb Z_p\times\mathbb Z_p$, $\cY=\mathbb Z_p$, and define the softened modular-addition target
\begin{equation}
  q_\beta(y\mid a,b)=
  \frac{\exp\{\beta\,\mathbf 1[y=a+b\;({\rm mod}\;p)]\}}
  {\exp(\beta)+p-1},
  \qquad \beta<\infty .
  \label{eq:qbeta}
\end{equation}
The lookup implementation assigns an independent quotient-logit vector to every pair $(a,b)$. The shared cyclic implementation uses logits that depend only on the residue $r=y-a-b\;({\rm mod}\;p)$.

This is the group-sharing theorem with $G=\mathbb Z_p^2$ acting by
\begin{equation}
  (s,t)\cdot(a,b,y)=(a+s,b+t,y+s+t).
\end{equation}
There is one orbit on inputs and $p$ orbits on input-output triples, classified by $r=y-a-b$. Hence:

\begin{corollary}[Same distribution, different local geometry]
\label{cor:modular-ranks}
For the softened modular-addition distribution in Eq.~\eqref{eq:qbeta}, after quotienting softmax gauges,
\begin{equation}
  r_{\rm lookup}=p^2(p-1),
  \qquad
  r_{\rm shared}=p-1.
\end{equation}
The quadratic active contributions (and the LLCs of these affine charts) are
\begin{equation}
  \lambda_{\rm lookup}^{\rm quad}=\frac{p^2(p-1)}2,
  \qquad
  \lambda_{\rm shared}^{\rm quad}=\frac{p-1}{2}.
\end{equation}
Both charts implement exactly the same conditional distribution at the base point.
\end{corollary}

The factor-$p^2$ gap is not a parameter-count artifact: by Proposition~\ref{prop:invariance}, adding dead parameters or duplicating the shared coordinates does not change the active rank. It is also not an invariant of the point distribution: by Remark~\ref{prop:ambient}, allowing arbitrary lookup perturbations around the shared point changes the local model and recovers the lookup rank.

Section~\ref{sec:operational} tests these ranks on trained checkpoints. Appendix~\ref{app:numerical-checks} gives separate deterministic algebra checks and distinguishes Fisher rank from hard-label score-Gram rank.

\section{Evaluating charts on neural networks}
\label{sec:operational}

A chart is a hypothesis about permitted interventions, not something predictions uniquely identify. Natural choices include perturbing tied convolutional kernels, varying an equivariant layer within its constrained parameter space, changing a compiled RASP program, or intervening on specified heads, MLP blocks, Fourier directions, or sparse-autoencoder features. These choices need external support and need not be unique.

\paragraph{Evaluation procedure.}
First specify the candidate and its intervention coordinates. Next measure its quotient-logit Jacobian $J$ over the evaluation domain and form $M=F_q^{1/2}J$. Count its independent directions, or inspect its spectrum when sharing is approximate. Finally ask whether those directions explain an independently specified intervention response: fit its tangent coefficients on one set of inputs and measure the Fisher-weighted residual on held-out inputs. A low residual supports the candidate relative to that response and intervention family. Rank alone is not a selection score: projecting onto a small subspace already limits rank by construction.

\subsection{Trained modular-addition checkpoints}
We train a two-hidden-layer tanh MLP of width 64 and a residue-shared network on the complete softened modular-addition domain ($\beta=3$). We measure the unrestricted chart, a prespecified shared projection of the same MLP Jacobian, and the shared architecture's own chart. All runs reach 100\% accuracy. Table~\ref{tab:trained-ranks} recovers the predicted ranks across three seeds for $p=5,7$ and two for $p=11$, at relative singular-value thresholds $10^{-6},10^{-8},10^{-10}$. Appendix~\ref{app:trained} gives the protocol and numerical fit errors.

\begin{table}[htbp]
\centering
\caption{Active ranks from trained checkpoints. Each entry is unchanged across the tested seeds and thresholds. The projected chart is imposed on the same unrestricted MLP; it is not discovered by training.}
\label{tab:trained-ranks}
\begin{tabular}{rrrrr}
\toprule
$p$ & Seeds & Unrestricted MLP & Shared projection & Shared architecture \\
\midrule
5 & 3 & 100 & 4 & 4 \\
7 & 3 & 294 & 6 & 6 \\
11 & 2 & 1210 & 10 & 10 \\
\midrule
Theory & & $p^2(p-1)$ & $p-1$ & $p-1$ \\
\bottomrule
\end{tabular}
\end{table}
\FloatBarrier

The interpretation is limited but useful: chart rank can be computed from a trained network, and the same checkpoint admits different chart ranks. Random-initialization controls at $p=5,7$ already have the same ranks (Appendix~\ref{app:trained}); the table therefore validates measurement, not emergence of modular structure. Fitting the function does not turn an unrestricted parameter chart into a shared chart.

\subsection{Controlled transformer candidate comparison}
To test discrimination rather than rank calculation, we compile direct and decomposed RASP programs for the adjacent-token rule ``previous token is a and current token is b'' using Tracr \citep{weiss2021thinking,lindner2023tracr}. They agree on all 98 output positions in the complete domain of 30 binary sequences of lengths one through four. At four compiler-known intermediate computations, each candidate independently changes the gains of two orthonormal residual directions. We compare the compiler pair with every incorrect named pair and 500 random pairs of the same dimension.

The target is the logit response to ablating the compiler pair. Candidate coefficients are fit on three sequence folds and evaluated on the fourth. The mean relative Fisher-weighted held-out residual is below $4.4\times10^{-16}$ for the compiler chart, at least $0.67$ for every named alternative, and $0.009$--$0.058$ for the best random pair in each task (Table~\ref{tab:tracr}). The defining ablations change 17.3\% of output positions. These results test the ability to explain a known causal response; the target itself uses compiler information.

\begin{table}[htbp]
\centering
\caption{Independent-gain chart comparison. Residuals are means over four held-out sequence folds; lower is better. Random columns summarize the fixed pool of 500 candidates per task, not repeated training runs.}
\label{tab:tracr}
\begin{tabular}{lrrrr}
\toprule
Intermediate & Compiler & Named best & Random best & Random median \\
\midrule
Direct: previous token & $0$ & 0.691 & 0.057 & 0.601 \\
Decomposed: previous token & $0$ & 0.691 & 0.030 & 0.606 \\
Decomposed: current-is-\texttt{b} & $1.1\times10^{-16}$ & 0.674 & 0.058 & 0.674 \\
Decomposed: previous-is-\texttt{a} & $4.4\times10^{-17}$ & 0.691 & 0.009 & 0.586 \\
\bottomrule
\end{tabular}
\end{table}
\FloatBarrier

\paragraph{Negative controls define the scope.}
At both final-output computations, all 500 random pairs tie the compiler chart within $10^{-6}$. Allowing arbitrary $2\times2$ mixing within each candidate subspace, rather than independent gains, also removes discrimination in three of four intermediate tasks. The positive result thus depends on the specified intervention family. It establishes neither prediction-only identification nor automatic circuit discovery in learned transformers.

\subsection{Approximate sharing: inspect the spectral tail}
\label{sec:approx}
Exact rank can jump when a small perturbation breaks symmetry. A more stable question is how much spectrum lies outside the shared directions. Fix parameter coordinates, let $A$ be the tangent map, and let $P_G$ be the $F_q$-orthogonal projection onto the shared subspace of dimension $r_G$. With $E=F_q^{1/2}(I-P_G)A$ and $H=A^\top F_qA$, Appendix~\ref{app:robustness} proves
\begin{equation}
  \lambda_{r_G+1}(H)\le \|E\|_{\rm op}^2,
  \qquad \sum_{i>r_G}\lambda_i(H)\le\|E\|_F^2,
  \label{eq:tail-bound-main}
\end{equation}
where eigenvalues are ordered decreasingly. Small symmetry-breaking sensitivity therefore implies a small tail, even if exact rank increases. The converse is not guaranteed: a small tail alone need not identify the correct shared subspace. Numerical almost-shared models confirm the bounds; they are spectral diagnostics in fixed coordinates, not coordinate-invariant estimates of the full LLC.

\section{Bayesian model selection and relation to SLT}
\label{sec:slt}

For a local statistical model with empirical loss $L_n$ and prior density positive near a minimizing set, SLT gives the free-energy expansion
\begin{equation}
  F_n=-\log\int \exp\{-nL_n(w)\}\varphi(w)\dd w
  = nL_n(w_*)+\lambda\log n-(m-1)\log\log n+O_p(1),
  \label{eq:slt-expansion}
\end{equation}
where $\lambda$ is the learning coefficient and $m$ is its multiplicity \citep{watanabe2009,watanabe2013,watanabe2018}. In an analytic regular rank-$r$ active chart with positive smooth prior, $\lambda=r/2$. Therefore, when two specified model-prior pairs fit the same distribution equally well and differ only in regular active rank, the lower-rank mechanism receives the smaller asymptotic complexity penalty. This is a local asymptotic comparison: prior mass, constants, and finite-sample effects can dominate at finite $n$.

\paragraph{What rank misses.}
For an analytic nonnegative loss germ with Hessian rank $r$, analytic splitting gives local coordinates in which $K=\|u\|^2+h(v)$, with $h\ge0$ having no quadratic part. Under a positive smooth local prior,
\begin{equation}
  \lambda(K)=r/2+\lambda(h)\ge r/2,
\end{equation}
where $\lambda(h)=0$ if $h$ is identically zero. Appendix~\ref{app:splitting} gives the argument. Thus rank measures the quadratic contribution, while higher-order directions can add complexity.

\begin{example}[Equal rank, different learning coefficients]
The losses $K_1(u,s)=u^2+s^4$ and $K_2(u,s)=u^2+s^6$ both have Hessian rank one. Nevertheless, their LLCs are $1/2+1/4=3/4$ and $1/2+1/6=2/3$ (Proposition~\ref{prop:protected-normal-form}). These behaviors also occur in finite-output charts: near a full-support distribution, logits varying as $(u,s^2)$ or $(u,s^3)$ have KL losses comparable to these forms. First-order mechanism equivalence does not imply equality of higher-order singularities.
\end{example}

The present results should be read as a local, finite-output complement to \emph{Programs as Singularities} \citep{murfetTroiani2025programs}. That work constructs smooth relaxations of Turing-machine codes and relates the Taylor expansion of the loss to error syndromes internal to a program. Here the ``code perturbations'' are mechanism charts in quotient logit space. The shared message is that Bayesian/SLT complexity can depend on implementation-local geometry even when predictive behavior agrees. The difference is that our main theorems are finite-dimensional linear and group-theoretic at second order; Appendix~\ref{app:protected} records a simple higher-order normal form that begins to recover the singular phenomena emphasized by SLT.

The theory also clarifies the relationship to flatness-based generalization arguments. Hessian spectra and flatness are not invariant under arbitrary neural reparameterizations \citep{dinh2017sharp}. Our active-rank statements deliberately avoid arbitrary parameter-space flatness: they are invariant only under mechanism-preserving changes that leave the tangent image in logit space unchanged. This is a weaker but better-posed claim.

\section{Related work}
\label{sec:related}

\paragraph{Classical Fisher geometry.} Lemma~\ref{thm:pullback} is closely related to standard Fisher-information geometry for parameterized statistical manifolds and tangent spaces \citep{amariNagaoka2000,kassVos1997,amari2016information}. An implementation chart is mathematically a local parameterization of a restricted statistical model; it does not replace that established concept. Our focus is the comparison of mechanism-supported restricted models through the same point distribution, after removing the per-input softmax gauge. Equality of tangent images separates a coordinate change from a change of perturbation family. The orbit-count formula then computes that dimension for exactly shared mechanisms. This comparison and computation, rather than a new notion of statistical manifold, are the contribution.

\paragraph{Singular learning theory and local complexity.} SLT replaces regular dimension penalties by birational invariants of the KL germ \citep{watanabe2009,watanabe2013,watanabe2018}; related singular-model and algebraic-statistics perspectives include \citet{drtonSturmfelsSullivant2009,drtonPlummer2017}. Earlier information-geometric and Bayesian complexity perspectives include \citet{balasubramanian1997occam,schwarz1978bic,mackay1992bayesian}. The LLC has recently been proposed as a practical singularity-aware complexity measure for deep networks \citep{lau2025llc}, with follow-up work connecting degeneracy to stagewise development and component specialization \citep{hoogland2024developmental,wang2024rllc}. Our theorems do not estimate LLCs in large networks; they identify a tractable class of implementation-local KL germs where the quadratic active contribution can be computed exactly.

\paragraph{Programs, algorithms, and structural priors.} Algorithmic information theory and MDL give function- or code-length-based accounts of simplicity \citep{solomonoff1964formal,rissanen1978modeling,hutter2005universal,vereshchagin2004kolmogorov}. Program-synthesis geometry and Programs as Singularities instead study smooth neighborhoods of program codes \citep{clift2021geometry,murfetTroiani2025programs}. Our contribution is closer to the latter: the relevant object is not only which function is computed, but which perturbations of the implementation are locally available.

\paragraph{Mechanistic interpretability and compiled transformers.} Mechanistic interpretability uses interventions, causal abstractions, and circuit-level hypotheses to describe internal algorithms \citep{elhage2021framework,geiger2023causal,chan2022causal}. RASP and Tracr provide programmatic laboratories where transformer mechanisms are known by construction \citep{weiss2021thinking,lindner2023tracr}. Benchmarks such as MIB emphasize that localization methods must be compared against strong patching and attribution baselines \citep{mueller2025mib}. Our paper supplies a theoretical target for one kind of diagnostic---mechanism-local dimension---rather than a new benchmark-winning localization algorithm.

\paragraph{Equivariance and sharing.} Weight sharing and equivariance are central tools for building models whose local perturbations respect a symmetry \citep{cohen2016group,kondor2018generalization,bronstein2021geometric}. Our finite-group theorem can be read as the local Fisher-rank consequence of exact sharing in the finite-output setting: invariant logit functions have one degree of freedom per input-output orbit, minus the input-gauge orbits.

\section{Limitations}
\label{sec:limitations}

The results require a finite evaluation domain and an externally specified chart. The trained-model experiment checks measurement; it does not show that sharing was learned. The Tracr experiment uses compiler-defined ablations and a restricted intervention family, with explicit non-identifiability controls. Neither experiment establishes chart discovery or performance on generic learned circuits. Evaluating complete Jacobians also becomes expensive as the domain and parameter count grow.

The orbit formula concerns exact sharing and full-support predictions. Approximate sharing admits the spectral bounds in Appendix~\ref{app:robustness}, but rank remains discontinuous and probabilities near zero make numerical rank ill-conditioned. The bounds depend on fixed intervention coordinates. Higher-order terms can change LLCs at identical active rank, and hard-label score-Gram ranks are not calibrated LLC estimates. Local asymptotic Bayesian complexity does not by itself predict finite-sample preference or training dynamics.

\section{Conclusion}
Equal predictions leave the allowed local changes undetermined. Specifying those changes makes a mechanism hypothesis measurable: finite-group sharing has an exact orbit-count dimension, and controlled experiments show how chart spectra and intervention residuals can be evaluated. The useful next step is evidence-based candidate generation in learned networks, with the same attention to negative controls and intervention choice.

{\small
\bibliographystyle{plainnat}
\bibliography{bibliography}
}

\appendix

\section{Proof of the logit-Fisher pullback lemma}
\label{app:pullback-proof}

Choose any representative $z_x\in\R^k$ of a quotient logit vector. The function
\begin{equation}
  \KL(q_x\|\softmax(z_x))
  = \sum_y q_x(y)\log q_x(y) - \sum_y q_x(y)z_x(y)
  + \log\sum_y \exp z_x(y)
\end{equation}
is invariant under $z_x\mapsto z_x+c\one$. Its gradient in $\R^k$ is $\softmax(z_x)-q_x$, and its Hessian is
\begin{equation}
  \diag(\softmax(z_x))-\softmax(z_x)\softmax(z_x)^\top.
\end{equation}
At $z_x^*=[\log q_x]$ this Hessian is $C_x=\diag(q_x)-q_xq_x^\top$. Since $q_x$ has full support, for every nonconstant representative $v$,
\begin{equation}
  v^\top C_x v = \operatorname{Var}_{Y\sim q_x}[v(Y)]>0,
\end{equation}
so $C_x$ is positive definite on $\R^k/\Span\{\one\}$. Summing with positive weights $\mu(x)$ gives the positive-definite quotient metric $F_q$ in Eq.~\eqref{eq:fisher-metric}.

Let $A=D\chi_A(0)$. Since $\nabla K(z^*)=0$, the second derivative chain rule has no term involving $D^2\chi_A(0)$, and therefore
\begin{equation}
  \nabla^2K_A(0)[u,v]=F_q(Au,Av).
\end{equation}
The nullspace is $\ker A$, because $F_q$ is positive definite. Hence the Hessian rank is $\dim\im A$. On the quotient $\R^{d_A}/\ker A$, this is a nondegenerate quadratic minimum. Under the analytic and prior assumptions stated in Section~\ref{sec:setup}, the zeta integral of a nondegenerate rank-$r$ quadratic form has leading pole at $-r/2$, giving quadratic active contribution $r/2$. If tangent-kernel coordinates are not removed or genuinely dead, this calculation describes only the active quadratic block.

\section{Proof of invariance and the perturbation-class observation}
\label{app:invariance-proof}

For Proposition~\ref{prop:invariance}, set $B=D\rho(0)$. The composed tangent map is $AB$. If $B$ is surjective, then $\im(AB)=\im A$, so the active rank is unchanged by Lemma~\ref{thm:pullback}. If $\rho$ is a local diffeomorphism, this is ordinary coordinate invariance. If $\rho(u,v)=u$, the extra $v$ coordinates do not enter the germ. If $\rho(u_1,u_2)=u_1+u_2$, the linear change of variables $(u_1,u_2)\mapsto (u_1+u_2,u_2)$ separates active and redundant coordinates.

For the learning-coefficient statement, assume the analytic and prior conditions in Proposition~\ref{prop:invariance}. The analytic constant-rank theorem gives local coordinates $(a,b)$ near $0\in V$ and $u$ near $0\in U_A$ in which $\rho(a,b)=a$ on the image coordinates. In these coordinates, the local zeta integral of the pullback germ is
\begin{equation}
  \int |K_A(\rho(a,b))|^z \varphi(a,b)\dd a\dd b
  =
  \int |K_A(a)|^z
  \left(\int \varphi(a,b)\dd b\right)\dd a ,
\end{equation}
over sufficiently small neighborhoods. The inner integral is a positive smooth factor with finite nonzero value near the origin, so it does not change the leading pole of the zeta function. Hence the RLCT, and therefore the learning coefficient, is unchanged. This is the formal version of saying that dead or duplicated coordinates contribute only finite fiber volume.

For Remark~\ref{prop:ambient}, the charts $z^*+S$ and $z^*+T$ have derivative images $S$ and $T$, respectively. Lemma~\ref{thm:pullback} gives ranks $\dim S$ and $\dim T$. This proves that changing the local perturbation family changes the statistical model being compared.

\section{Proof of the group-sharing theorem}
\label{app:group-proof}

Let $V=\R^{\cX\times\cY}$ be raw logits and let $B\subset V$ be the gauge subspace of functions constant over $y$ for each fixed $x$. Then $\cL=V/B$. Compatibility of the $G$ action with the projection $\cX\times\cY\to\cX$ implies $B$ is $G$-stable.

Because $G$ is finite and we work over $\R$, the Reynolds operator $R(v)=|G|^{-1}\sum_{g\in G}gv$ projects every representation onto its invariant subspace. Hence taking invariants is an exact functor, and the short exact sequence $0\to B\to V\to V/B\to0$ yields
\begin{equation}
  0\to B^G\to V^G\to (V/B)^G\to0.
\end{equation}
Thus $\dim\cL^G=\dim V^G-\dim B^G$.

A function on a finite $G$-set is invariant exactly when it is constant on orbits. Therefore $\dim V^G=\#((\cX\times\cY)/G)$. Similarly, a gauge function is specified by one scalar per input, and $G$-invariance means this scalar is constant on input orbits; hence $\dim B^G=\#(\cX/G)$. This proves Eq.~\eqref{eq:orbit-rank}. Since the target is full support, Lemma~\ref{thm:pullback} converts this tangent dimension into active Hessian rank.

\section{Higher-order protected directions}
\label{app:protected}

The pullback rank is a second-order invariant. SLT is sensitive to higher-order terms when some first-order perturbations are protected. The following normal form records the basic calculation used throughout singular learning theory.

\begin{proposition}[Product normal form]
\label{prop:protected-normal-form}
Suppose an analytic nonnegative germ is locally comparable to
\begin{equation}
  f(u,s,t)=\sum_{i=1}^{r}u_i^2+\sum_{j=1}^{m}s_j^{2a_j},
  \qquad a_j\in\N,
  \label{eq:product-normal-form}
\end{equation}
where $t$ are dead coordinates. That is, $c f\le K\le C f$ near the origin for some $0<c<C<\infty$. Then $K$ has the same local learning coefficient as $f$,
\begin{equation}
  \lambda(K)=\frac r2+\sum_{j=1}^{m}\frac{1}{2a_j}.
\end{equation}
\end{proposition}

\begin{proof}
For a nonnegative analytic germ with positive smooth local density, the learning coefficient is the critical integrability exponent: $\int K^{-c}$ is finite for $c<\lambda$ and divergent for $c>\lambda$. The inequalities $c_0f\le K\le C_0f$ give the same threshold for $K$ and $f$.

To compute it for the displayed sum, substitute $u_i=\epsilon^{1/2}\widetilde u_i$ and $s_j=\epsilon^{1/(2a_j)}\widetilde s_j$ in its sublevel set $\{f\le\epsilon\}$. The rescaled active set has finite nonzero volume independent of $\epsilon$. Its Jacobian contributes $\epsilon^{r/2+\sum_j1/(2a_j)}$; dead coordinates supply a finite factor. A positive smooth prior is bounded above and below locally and leaves this exponent unchanged. This sublevel exponent equals the critical integrability threshold, proving the formula \citep{watanabe2009,watanabe2018}.
\end{proof}

This proposition explains the difference between dead code, duplicated code, and error-protected code. Dead coordinates do not affect $K$ and contribute nothing. Duplicated coordinates such as $(u_1+u_2)^2$ become one quadratic coordinate plus one dead coordinate. A protected coordinate that first changes logits at order $a>1$ contributes $1/(2a)$ rather than $1/2$. Programs as Singularities develops a much richer version of this idea for noisy Turing-machine codes and error syndromes \citep{murfetTroiani2025programs}.

\section{Modular-addition derivations}
\label{app:modular-derivations}

Let $q=q_\beta$ in residue coordinates, with
\begin{equation}
  q_0=\frac{e^\beta}{e^\beta+p-1},
  \qquad
  q_1=\frac{1}{e^\beta+p-1}.
\end{equation}
The per-input Fisher matrix is $C=\diag(q)-qq^\top$. It has one gauge eigenvector $\one$ with eigenvalue zero. Among incorrect residues, the $p-2$ independent contrasts have eigenvalue $q_1$. The remaining centered direction contrasts the correct residue with the average incorrect residue and has eigenvalue $p q_0q_1$. Thus $\rank C=p-1$ on the quotient.

In the lookup chart, the expected KL Hessian is block diagonal with $p^2$ copies of $C/p^2$, one for each input pair. Hence the rank is $p^2(p-1)$. In the shared cyclic chart, every input uses the same residue-logit perturbation, relabeled by the cyclic output action, so input averaging gives one copy of $C$ and rank $p-1$.

For the hard-label influence diagnostic, the score vector at the correct label is $h=q-e_0\ne0$. Lookup scores occupy disjoint blocks, so the input-level Gram matrix is $\|h\|^2I_{p^2}$. Shared cyclic scores all equal $h$, so the Gram matrix is $\|h\|^2\one\one^\top$.

\FloatBarrier
\section{Deterministic numerical checks and score conventions}
\label{app:numerical-checks}

The accompanying code supplement contains finite-orbit enumeration, explicit small-matrix Hessian checks, per-input Fisher spectra, quadratic-posterior calculations, and protected/dead/duplicated normal-form checks. These verify algebra, separately from the trained and compiled-network experiments. Running instructions, environment versions, and saved results are included in the supplement. The original anonymous artifact remains at \url{https://anonymous.4open.science/r/review-artifact-2026-16-FD3C}; it predates the rebuttal experiments and is not their reproducibility package.

\begin{table}[htbp]
\centering
\small
\caption{Deterministic modular-addition ranks for the lookup and shared cyclic charts. The values are independent of the finite temperature $\beta$; the supplement recomputes them by explicit small-matrix checks for these $p$ values.}
\label{tab:modular-ranks}
\vspace{0.35em}
\begin{tabular}{rrrrrr}
\toprule
$p$ & $r_{\rm lookup}$ & $r_{\rm shared}$ & $\lambda_{\rm lookup}^{\rm quad}$ & $\lambda_{\rm shared}^{\rm quad}$ & influence ranks \\
\midrule
3  & 18   & 2  & 9   & 1 & $9/1$ \\
5  & 100  & 4  & 50  & 2 & $25/1$ \\
7  & 294  & 6  & 147 & 3 & $49/1$ \\
11 & 1210 & 10 & 605 & 5 & $121/1$ \\
\bottomrule
\end{tabular}
\vspace{0.35em}
\end{table}
\FloatBarrier

These are deterministic algebraic sanity checks, not empirical validation. Their purpose is to make the size of the rank gap visible and to catch implementation mistakes in the finite calculations.

\paragraph{Influence-rank diagnostic.} If per-input scores are taken from the hard-label cross-entropy at the matched soft distribution, the same contrast appears in an observable Gram matrix. Let the label be $y=a+b$ and let $h=q_\beta- e_0$ in residue coordinates. Lookup scores for different inputs lie in disjoint parameter blocks, so the input-level Gram matrix is $\|h\|^2I_{p^2}$ and has rank $p^2$. Shared cyclic scores are identical across inputs, so the Gram matrix is $\|h\|^2\one\one^\top$ and has rank $1$. This diagnostic convention is separate from the Fisher rank: the population sampled-label Fisher score covariance recovers the Hessian rank, while finite empirical score matrices require enough sampled labels to span the quotient directions and soft-target KL gradients vanish at the optimum.

\section{Approximate sharing and conditioning}
\label{app:robustness}

\begin{proposition}[Spectral tail outside a candidate shared subspace]
Fix a full-support conditional distribution, positive input masses, and Euclidean coordinates on the chart parameters. Let $F_q$ denote the positive-definite quotient-logit metric, $A$ a tangent map, and $S$ a subspace of dimension $r$. Let $P$ be the $F_q$-orthogonal projection onto $S$ and set $M=F_q^{1/2}A$, $B=F_q^{1/2}PA$, and $E=M-B$. With singular values and eigenvalues ordered decreasingly (and padded by zeros),
\begin{align}
 \sigma_{r+1}(M)&\le\|E\|_{\rm op},\\
 \lambda_{r+1}(A^\top F_qA)&\le\|E\|_{\rm op}^2,\\
 \sum_{i>r}\lambda_i(A^\top F_qA)&\le\|E\|_F^2.
\end{align}
The statement applies in particular to $S=\cL^G$ with $r=r_G$ from Theorem~\ref{thm:group-sharing}.
\end{proposition}

\begin{proof}
Since $\im(PA)\subseteq S$, $B$ has rank at most $r$. The variational characterization of singular values bounds the $(r+1)$st singular value of $M$ by its operator-norm distance to any rank-at-most-$r$ matrix, hence by $\|M-B\|_{\rm op}$. For the Frobenius bound, take a singular value decomposition of $M$. Truncating it after $r$ terms minimizes squared Frobenius approximation error over rank-at-most-$r$ matrices; the minimum is $\sum_{i>r}\sigma_i(M)^2$ and cannot exceed $\|M-B\|_F^2$. Finally, $A^\top F_qA=M^\top M$, so its eigenvalues are the squared singular values of $M$.
\end{proof}

The proof requires a candidate subspace, not exact equivariance of the base distribution. When $q$ is not equivariant, an affine chart through its base logits with tangent $\cL^G$ represents shared \emph{perturbations}, not an exactly equivariant predictive family. The $F_q$-orthogonal projection remains well-defined, but need not equal unweighted group averaging. At a checkpoint that does not exactly fit the chosen target, $J^\top F_qJ$ is a target-weighted tangent diagnostic; it need not equal the exact loss Hessian there, whose chain rule also includes a nonzero-gradient term.

\paragraph{Full-support conditioning.}
For one input, choose the representative satisfying $q_x^\top v=0$. If $q_x(y)\ge\alpha>0$ for every output, then
\begin{equation}
 \alpha\|v\|_2^2\le v^\top C_xv
 =\sum_yq_x(y)v_y^2\le\|v\|_2^2.
\end{equation}
This follows directly from the diagonal weights after the centering term vanishes. Thus Fisher weighting is nondegenerate away from the simplex boundary. Numerical rank still requires a singular-value gap relative to the threshold; full support alone does not supply that gap. Small probabilities can make nonzero directions difficult to resolve even though the exact orbit-count rank is unchanged.

\paragraph{Numerical checks.}
The supplement tests $p=5,7$ at $\beta=3$. Synthetic shared charts receive symmetry-breaking corrections with weighted operator norms $10^{-1},10^{-2},10^{-3}$. Almost-shared networks add a $\gamma$-scaled dense correction to a residue-shared network, with $\gamma\in\{10^{-1},10^{-2}\}$ and seed zero. Every recorded operator and Frobenius bound holds. For $p=7$, $\gamma=10^{-2}$, the weighted residual norm is approximately $3.7\times10^{-3}$, the sixth eigenvalue is $0.15$, and the seventh is $1.4\times10^{-5}$, approximately the squared bound. These are controlled numerical checks of the inequality, not a test of spontaneous learned equivariance. The parameter coordinates and their scaling are fixed throughout.

\section{Quadratic and higher-order contributions to the LLC}
\label{app:splitting}

Let $K$ be a nonnegative analytic germ with $K(0)=0$ and Hessian rank $r$. Split its initial coordinates into $u$ in the positive Hessian block and $v$ in its kernel. The analytic implicit-function theorem solves $\partial_uK=0$ as $u=\psi(v)$ near zero. Set $h(v)=K(\psi(v),v)$. The analytic Morse lemma with parameters then gives a local analytic coordinate change for which
\begin{equation}
 K(u,v)=\sum_{i=1}^r u_i^2+h(v).
\end{equation}
Nonnegativity implies $h\ge0$; the kernel choice gives $D^2h(0)=0$. A positive smooth local prior and the coordinate-change Jacobian are bounded above and below by positive constants on a sufficiently small neighborhood, so they do not affect the leading integrability exponent.

For the displayed separated form, the local Laplace integral with constant density factors into a Gaussian integral in $u$ and an integral of $e^{-nh(v)}$ in $v$. The Gaussian factor has leading order $n^{-r/2}$. The analytic nonnegative germ $h$ contributes $n^{-\lambda(h)}$ up to logarithmic factors by the usual local SLT asymptotics \citep{watanabe2009,watanabe2018}. Comparing with the original positive density yields
\begin{equation}
 \lambda(K)=r/2+\lambda(h).
\end{equation}
If $h\equiv0$, its integral has constant positive volume and we set $\lambda(h)=0$. This establishes the lower bound and its limitation: a tangent-kernel direction can be inactive quadratically without being dead at higher orders. The statement is local to the chosen germ and positive prior, not a formula obtained by averaging ranks over training checkpoints.

\section{Trained-checkpoint protocol and initialization controls}
\label{app:trained}

\paragraph{Data and models.}
For each $p\in\{5,7,11\}$, we enumerate all $p^2$ pairs $(a,b)$ with uniform input mass and target $q_\beta$ from Eq.~\eqref{eq:qbeta}, using $\beta=3$. The unrestricted network takes concatenated one-hot encodings of $a,b$ (dimension $2p$), has two tanh hidden layers of width 64, and outputs $p$ logits. The shared network maps a one-hot encoding of residue $y-a-b$ through a width-16 tanh hidden layer and one scalar output; the same map is reused for every $(a,b,y)$. Both include biases. Thus sharing is architectural in the second model and imposed as a candidate intervention in the projected first model.

\paragraph{Training.}
We minimize full-domain soft-target cross-entropy in float64 with PyTorch L-BFGS, learning rate 1, at most 500 iterations, strong-Wolfe line search, gradient tolerance $10^{-12}$, and objective-change tolerance $10^{-14}$. Seeds are $0,1,2$ for $p=5,7$ and $0,1$ for $p=11$. There is no train/test split in this fitting experiment: it is a complete-domain geometry check, not a generalization benchmark. Widths and optimizer settings are fixed across seeds; no best-seed selection is used.

\paragraph{Coordinates and projection.}
The quotient coordinates are $z_y-z_{p-1}$ for $y=0,\ldots,p-2$. Their metric uses the upper-left $(p-1)\times(p-1)$ block of $C_x$, scaled by $1/p^2$. We differentiate these coordinates with respect to all parameters to form $J$. For sharing, let $S$ contain the quotient-coordinate effects of the $p-1$ residue-kernel contrasts. If $Q$ is an orthonormal basis of $\im(F_q^{1/2}S)$, the projected spectrum is computed from $Q^\top F_q^{1/2}J$. This is the nonzero spectrum of the Fisher-orthogonal projected tangent, expressed without an explicit large projector. The original rebuttal script used a Euclidean quotient-coordinate projection; the present implementation uses Fisher-orthogonal projection consistently. Independent reruns reproduce every reported rank with this convention.

\paragraph{Numerical precision and variability.}
A singular value counts as nonzero when it exceeds $t$ times the largest, for $t\in\{10^{-6},10^{-8},10^{-10}\}$. All seeds and thresholds give the entries of Table~\ref{tab:trained-ranks}; the observed rank range for each entry is a singleton. Every trained model has 100\% full-domain classification accuracy. In the camera-ready rerun, mean KL is below $4\times10^{-12}$ for $p=5,7$ and $1.4\times10^{-9}$ for $p=11$. The archived rebuttal run had a smaller $p=11$ maximum ($2.6\times10^{-10}$); numerical optimizer stopping changes this fit error without changing any rank. These finite-precision fits approximate, rather than exactly satisfy, the theorem's matching-distribution hypothesis.

\paragraph{Random-initialization control.}
With training disabled and the same three seeds for $p=5,7$, the unrestricted, projected, and shared ranks are respectively $100/4/4$ and $294/6/6$ at all three thresholds. At initialization the target-weighted pullback is not generally the loss Hessian, since the target has not been fitted. This control shows that the rank gap can already follow from the architecture and candidate choice. It rules out interpreting the trained rank table as evidence that modular structure emerged during optimization.

\section{Compiled-transformer protocol and full controls}
\label{app:tracr}

The direct RASP program detects an adjacent \texttt{ab} pair through a joint rule. The decomposed program combines separate previous-is-\texttt{a} and current-is-\texttt{b} predicates. Tracr compiles both to transformer weights. We enumerate all 30 sequences over $\{\texttt{a},\texttt{b}\}$ of lengths one through four, prepend the compiler's beginning-of-sequence token, and score the 98 non-BOS positions. The two programs have zero mismatched positions. The intervention-capable forward pass is checked against the compiled model before candidate scoring.

For a component output row $h$ and orthonormal residual directions $u_1,u_2$, a candidate's independent-gain chart is
\begin{equation}
 h\longmapsto h+\sum_{i=1}^2\alpha_i(hu_i)u_i^\top.
\end{equation}
The compiler supplies a pair of categorical features at each selected component. Setting both gains to $-1$ ablates that pair and defines the target logit-contrast response $b$. Every incorrect pair of named residual directions and 500 random orthonormal pairs is evaluated, using seed 20260726 and the same two-coordinate family. Candidate tangents $J_U$ are derivatives at zero intervention, computed by automatic differentiation through the downstream model. The finite ablation response is not assumed to be linear a priori; the residual measures how well a tangent predicts it.

We use the softened binary reference distribution with $\beta=3$. Its binary quotient-logit Fisher weight is the same scalar $q_0q_1$ at every position, with uniform position mass $1/98$. Therefore Fisher weighting cancels from the relative residual in this particular benchmark; discrimination comes from the response directions, not heterogeneous confidence weights.

Sequences are enumerated deterministically and assigned to four folds by sequence index modulo four. For each held-out fold $f$, least squares on the other folds gives
\begin{equation}
 \widehat\alpha_{-f}=\arg\min_\alpha\|W_{-f}(b_{-f}-J_{U,-f}\alpha)\|_2^2,
 \qquad
 R_f=\frac{\|W_f(b_f-J_{U,f}\widehat\alpha_{-f})\|_2}{\|W_fb_f\|_2}.
\end{equation}
Here $W=F_q^{1/2}$ in binary contrast coordinates, and the pseudoinverse cutoff is $10^{-10}$. Rows from one sequence stay together. We report the arithmetic mean of the four valid fold residuals, not a test-set discovery accuracy. The candidate pool is fixed before residual scoring; comparisons use a practical tie tolerance of $10^{-6}$.

All four compiler charts have rank two and are uniquely best among the named pairs; none of the 500 random candidates per task ties within tolerance. The defining ablation changes 17 of 98 output positions in each task. These descriptive comparisons do not establish a distribution-free significance guarantee for circuit recovery. The intervention target is compiler-supported ground truth, and the sequence domain is deliberately small.

\paragraph{Output and broader-intervention controls.}
For both final-output components, all 500 random independent-gain pairs tie the compiler chart. The broader family $h\mapsto h+(hU)MU^\top$ permits four coordinates (all entries of a $2\times2$ matrix $M$), with the same family used for every candidate in that comparison. In the two previous-token tasks and the previous-is-\texttt{a} task, 99.8--100\% of random subspaces tie. Current-is-\texttt{b} remains separated (best random residual about $0.006$). Independent gains distinguish individual directions; arbitrary mixing depends only on their span. Broadening the family thus changes the mechanism hypothesis and can erase identifiability.

\paragraph{Reproduction and compute.}
The supplement includes the exact Tracr source snapshot used for compilation, its Apache-2.0 license, checksums, package versions, benchmark script, per-candidate scores, and summary tables. The camera-ready verification uses Python 3.11, NumPy 1.26.4, JAX/JAXlib 0.4.28, Haiku 0.0.12, Chex 0.1.86, and PyTorch 2.9.0 for the separate modular experiments. All reported checks run on an AMD Ryzen 9 7900X3D CPU, with one BLAS/OpenMP thread; no external dataset is needed. Verification takes approximately 5 seconds for the six $p=5,7$ fits, 16 seconds for the two $p=11$ fits, 2 seconds for initialization controls, 11 seconds for Tracr, and 3 seconds for approximate-sharing checks. Peak resident memory is below 1 GiB per process. Exact measurements and commands are recorded with the supplement. These measurements cover verification of the reported experiments, not the broader exploratory research preceding this paper.

\end{document}